\documentclass[]{spie}  

\usepackage{amsfonts}
\usepackage{graphicx}
\usepackage[colorlinks=true, allcolors=black]{hyperref}
\usepackage{amsmath}
\usepackage{amssymb}
\usepackage{booktabs}
\usepackage{multirow}
\usepackage[normalem]{ulem}
\useunder{\uline}{\ul}{}
\usepackage{algorithm}
\usepackage{algorithmic}
\usepackage{bbm}
\usepackage{adjustbox}
\usepackage{wrapfig}
\usepackage{times}
\title{\Large\textbf{Confidence-Aware Calibration and Scoring Functions for Curriculum Learning}}

\author[]{Shuang Ao, Stefan Rueger, Advaith Siddharthan}
\affil[]{Knowledge Media Institute, The Open University, UK}

\begin{document} 
\maketitle

\begin{abstract}

Despite the great success of state-of-the-art deep neural networks, several studies have reported models to be over-confident in predictions, indicating miscalibration. Label Smoothing has been proposed as a solution to the over-confidence problem and works by softening hard targets during training, typically by distributing part of the probability mass from a `one-hot' label uniformly to all other labels. However, neither model nor human confidence in a label are likely to be uniformly distributed in this manner, with some labels more likely to be confused than others. In this paper we integrate notions of  model confidence and human confidence with label smoothing, respectively \textit{Model Confidence LS} and \textit{Human Confidence LS}, to achieve better model calibration and generalization. To enhance model generalization, we show how our model and human confidence scores can be successfully applied to curriculum learning, a training strategy inspired by learning of `easier to harder' tasks. A higher model or human confidence score indicates a more recognisable and therefore easier sample, and can therefore be used as a scoring function to rank samples in curriculum learning. We evaluate our proposed methods with four state-of-the-art architectures for image and text classification task, using datasets with multi-rater label annotations by humans. We report that integrating model or human confidence information in label smoothing and curriculum learning improves both model performance and model calibration. The code are available at \url{https://github.com/AoShuang92/Confidence_Calibration_CL}.

\end{abstract}
\section{Introduction}
\label{sec:intro}
State-of-the-art models have achieved impressive performance in classification tasks, comparable to and sometimes even better than human judgment. Although models have achieved high accuracy, they can be poorly calibrated and over-confident in their predictions, leading to untrustworthiness, especially in sensitive applications. Label Smoothing (LS) is a regularization~\cite{szegedy2016rethinking} or confidence calibration~\cite{muller2019does} technique that spreads the one-hot label into a typically uniformly distributed soft-label. However, in terms of model learning capacity, the confidence scores of the trained model are not uniformly distributed. Hence, there is inconsistency between the uniform soft-labels introduced by LS and the prediction confidence of the model. On the other hand, recent work with the CIFAR10-H dataset~\cite{peterson2019human} observes that the  human uncertainty or confidence in labels, observed through crowdsourced annotations, has significant influence on model generalization. Similar to model confidence, human confidence is also not uniformly distributed, which motivates our investigation of Label smoothing methods that are aware of both model and human label confidence with respect to model generalization and calibration.

Curriculum learning (CL)~\cite{bengio2009curriculum}, inspired by human and animal learning, is a training strategy for deep neural networks that progresses from easy to hard tasks or samples. Recent studies have shown that
CL improves model generalization for various tasks, such as computer vision~\cite{guo2018curriculumnet,wei2021learn}, natural language understanding~\cite{xu2020curriculum,tay2019simple}, and reinforcement learning~\cite{ren2018self}. One of the critical factors for successful CL framework design is accurately ranking samples in terms of difficulty level. There are various methods designed to set the ranking threshold for CL, such as
sentence length~\cite{platanios2019competence} and word frequency~\cite{kocmi2017curriculum} for linguistic input, and number of objects~\cite{wei2016stc} and boundary information~\cite{qin2020balanced} for visual input. Recently, the model confidence score~\cite{penha2020curriculum,zhang2018empirical} has been used to rank training samples for CL, which is obtained from the maximum probability or predicted class probability of the model output. We argue that the softmax confidence for an incorrect label does not accurately represent the model confidence for this sample. Hence, the predicted probability of the correct target label should be taken as the model confidence score to determine the difficulty level for a sample.

Human annotation has also been utilized to rank samples in the CL framework. For instance, in the PASCAL VOC 2012 dataset~\cite{everinghampascal}, annotator response times in the visual search task were converted to image difficulty scores, which were further adapted by Ionescu et.al~\cite{tudor2016hard} to rank images for CL. The difficulty level of training samples can also be measured directly from human observation. For example, raters were directly asked to rank images in the training set from easiest to hardest to build the CL framework~\cite{pentina2015curriculum}. We instead infer difficulty from multi-rater annotations, based on the level of agreement of the raters on that sample. For instance, a image annotated by 10 annotators in the CIFAR10-H dataset~\cite{peterson2019human} as `bird' eight times and as `airplane' two times can be considered easier than another image which is annotated as each five times. 

In this paper, we investigate the impact of both model and human confidence in model calibration and curriculum learning strategies. \textit{Model confidence} ($M_c$) is the predicted probability for the target label, which we pre-compute using an independent baseline model. 
\textit{Human confidence} ($H_c$) is derived from the annotation distribution for a label, which reflects human perception of the samples. To better calibrate the model, we use $M_c$ and $H_c$ to replace the uniformly distributed soft-labels used by standard LS. In terms of the CL framework, a higher $M_c$ indicates the stronger prediction of a model, i.e. the sample is easier with respect to the learning capacity of model. Similarly, a sample with less disagreement between raters is easier to recognize. Therefore, $M_c$ and $H_c$ can be used to set the difficulty level for samples for CL. Our contributions and findings in this work are summarized as follows:
\begin{enumerate}
\item We improve upon the uniform distribution soft-label in Label Smoothing by using model and human confidence to distribute label probability more intelligently;

\item We propose a novel method to apply model and human confidence to rank samples for curriculum learning;

\item Our empirical studies show that:
\begin{enumerate}
\item  Confidence-aware calibration and scoring functions for curriculum learning outperform conventional loss functions;
 \item Human confidence provides slightly better guidance in effective learning than model confidence, but requires additional costs for data annotation;
 \item Fusing model and human confidence is not as effective as expected, and performs worse than their individual use.
 \end{enumerate}
\end{enumerate}

\section{Related Work}
\label{sec:related work}
\paragraph{Label Smoothing.}
LS~\cite{szegedy2016rethinking} is a strategy to regularize the network to reduce over-confidence and miscalibration by computing the cross-entropy loss with uniformly squeezed labels instead of one-hot labels. It has been successfully applied in many state-of-the-art deep learning models in the training process with a range of tasks, such as image classification~\cite{zoph2018learning}, speech recognition~\cite{chorowski2016towards} and machine translation~\cite{vaswani2017attention}. LS is effective and used widely, and yields better model calibration and confidence predictions~\cite{muller2019does}.  It demonstrates better feature representations by tightening within a cluster but enlarging the differences across clusters~\cite{vaswani2017attention}. 

As a regularization technique for model calibration, label smoothing has also been applied to learning from noisy labels. Recent deep learning models are easily over-fitting with noisy labels~\cite{zhang2021understanding}, and training with label smoothing can significantly improve the model performance under various levels of noise~\cite{lukasik2020does}. Label smoothing has thus been shown  to improve the model performance and uncertainty estimates for model learning from both clean and noisy labels. 

Label smoothing is integrated with cross-entropy loss as follows. Suppose $p_n$ is the true label and $\hat{p}_n$ is the predicted probability of the $n^{th}$ class. For a network trained with one-hot label, the minimized cross-entropy value ($CE$) between the target and prediction is:
\begin{equation}
\small
\label{eq:ce}
CE(p,\hat{p} ) = -\sum_{n=1}^{N}p_n\log(\hat{p}_n)
\end{equation}
Using uniform label smoothing to soften the one-hot label with the parameter $\alpha$, the soft label is: 
\begin{equation}
\small
\label{eq:ls}
p^{LS}_n = p_n(1-\alpha) + \alpha/N 
\end{equation}

Then, for the model trained with uniform soft labels, the cross-entropy loss is: 
\begin{equation}
\small
\label{eq:cels}
CE(p,\hat{p})^{LS} = -\sum_{n=1}^{N}p^{LS}_n\log(\hat{p}_n) 
\end{equation}

In this work we extend LS to non-uniformly distributed soft labels.

\paragraph{Curriculum Learning.}
Bengio et al. \cite{bengio2009curriculum} showed in their original work that curriculum learning (CL), an `easy to hard' training strategy for machine learning models, performs  better than training with randomly presented samples. More specifically, CL starts from the easier data, then gradually increases the complexity of data until the training has used the whole dataset. CL has been widely applied in recent state-of-the-art deep neural networks and has been shown to benefit model generalization in various applications, including histopathology image classification in the medical field~\cite{wei2021learn}, contextual difficulty generator for long narratives in natural language understanding~\cite{tay2019simple},
multi-modality data synchronization with self-supervised learning~\cite{korbar2018cooperative}, and for generative adversarial networks~\cite{ghasedi2019balanced}. 

One key issue in designing a Curriculum Learning (CL) framework is to determine how to rank training samples accurately in terms of difficulty level. Suppose in a training set ${D}$, $x$ represents the sample and $y$ is the corresponding true label. $S$ is the scoring function that creates subsets in the training set with the ranking threshold $\mu$. The $\mu$ threshold assigns the learning order to training subsets, such that subsets which are used for training earlier are easier for the model to learn. Each subset $d\{x, y\}$ can be represented as 
$d\{x, y\}=\mathcal{S}(D\{x, y\}, \mu)$. 

In this work, instead of scoring subsets of training samples, we utilize curriculum criteria to calculate loss from easier to harder tasks over the training epochs.

\paragraph{Human Uncertainty.}
 When human judgements are subjective, datasets should contain multiple human judgments for samples, to reflect the distribution of responses possible. Such datasets are in reality hard to come by in sufficient scale to deploy neural models. The three we are aware of wer created through crowdsourcing platforms such as Amazon Mechanical Turk~\cite{buhrmester2016amazon}, where many annotators are asked to label the same data into one or more classes. For instance, the ArtEmis dataset~\cite{achlioptas2021artemis} contains the human annotation of visual art with 9 emotion classes; the CIFAR10-H dataset~\cite{peterson2019human} includes 10,000 images rated by crowdsourcing with 10 object classes; and the WikiArt dataset~\cite{mohammad2018wikiart} contains rating for both image and text into 20 emotion classes. These datasets have been used to train deep learning models and have proved beneficial for model generalization and performance. For example, the captioning system built on the ArtEmis dataset is impressive in revealing the semantic and abstract content of images~\cite{peterson2019human}. Models trained with human uncertainty have previously also proved to be more robust under adversarial attack~\cite{peterson2019human}.

\section{Methodology}
\label{sec:method}

In this section, we describe how we integrate model and human confidence into Label Smoothing to better calibrate the model and to design the curriculum learning framework.

\subsection{Model and Human Confidence}

The output probability of a model denotes its confidence in the predicted class. In this paper, instead of using the output probability for the \emph{predicted} class, we use the output probability for the \emph{ground truth} for the model confidence ($M_c$). These are pre-computed using a baseline model with standard independent identically distributed ($iid$) training. 

We compute the human confidence for a class ($H_c$) from the standard deviation of the full-label distribution~\cite{peterson2019human}.

\begin{figure}[!h]
\centerline{\includegraphics[width=1\textwidth]{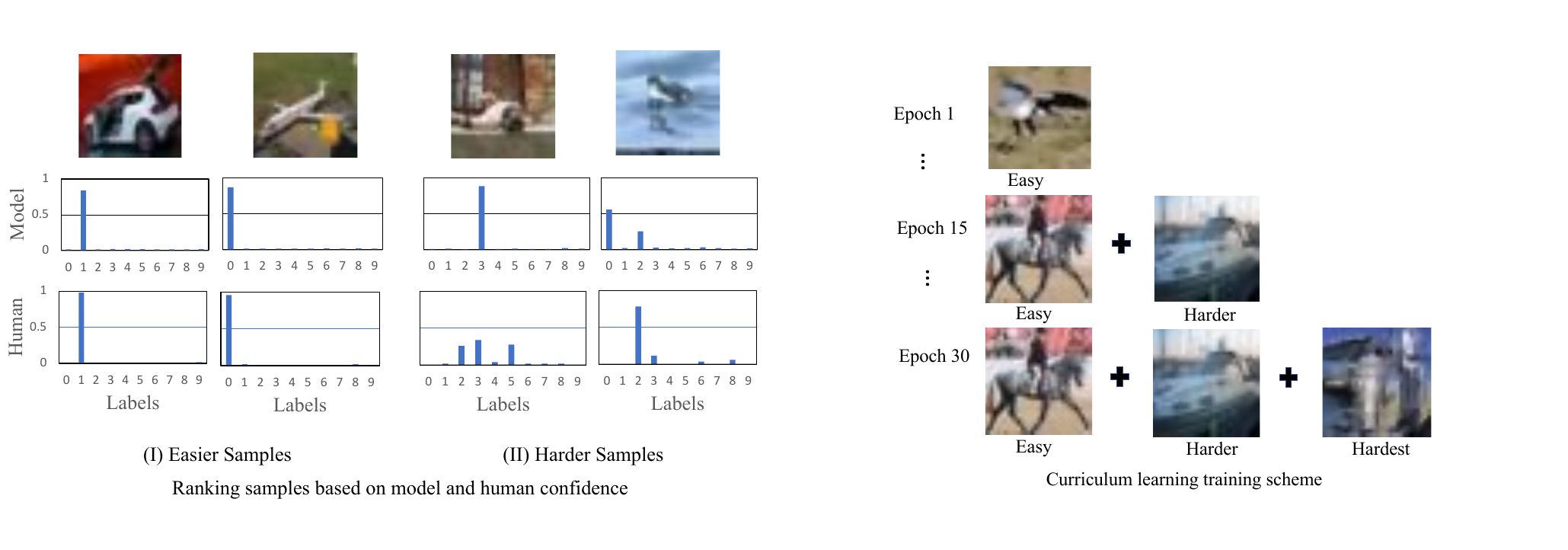}}
\caption{Left:Integrating confidence into label smoothing. Model and human confidence indicating easier and harder samples from CIFAR10-H dataset. Right: Curriculum learning training scheme with model and human confidence as the ranking criteria for samples from CIFAR10-H dataset.}
\label{fig:scoring ls}
\end{figure}

Figure~\ref{fig:scoring ls} (left) visually illustrates the model and human confidence of images. Higher confidence indicates a stronger judgement from the model or humans towards certain targets, suggesting easier tasks. 

\subsection{Non-uniform Label Smoothing using Machine and Human Confidence}

The smoothing factor $\alpha$ in LS (see Eq. \ref{eq:ls}) reweights the one-hot label into a uniformly distributed soft-label. Overlooking the model confidence in prediction and human confidence in the target leads to inconsistency in model learning. To tackle this issue, we incorporate model and human confidence with LS. 

\paragraph{Model Confidence Label Smoothing ($M_c LS$).}
Suppose $m = (m_1, m_2,..., m_N)$ is the model confidence for a target distribution with $N$ classes, then the $M_c LS$ smoothing factor $\alpha^{m} \in \mathbb{R}^{N}$ is formulated as:
\begin{equation}
\small
\label{eq:HCLS}
\alpha^{m} = \alpha + \gamma m 
\end{equation}

where $\gamma$ is the weighting parameter to control the effect of average confidence score into smoothing factor $\alpha$. Based on Eq.~\eqref{eq:ls} of LS, the true label of $n^{th}$ class $p_n$ with $M_c LS$ is:

\begin{equation}
\small
\label{eq:MCLS}
p^{\it M_c LS}_n = p_n\cdot(1-\alpha^{m}) + \alpha^{m}/N
\end{equation}

\paragraph{Human Confidence Label Smoothing ($H_c LS$).}
Suppose $h=(h_1, h_2, ..., h_n)$ is the human confidence for a target distribution with $N$ classes, then the $H_c LS$ smoothing factor $\alpha^{h}\in \mathbb{R}^{N}$ is written as:

\begin{equation}
\small
\label{eq:HCLS}
\alpha^{h} = \alpha + \gamma h
\end{equation}

According to Equation~\eqref{eq:ls}, the true label of the $n^{th}$ class $p_n$ is as follows:
\begin{equation}
\small
\label{eq:HCLS}
p^{\it H_c LS}_n = p_n\cdot(1-\alpha_n^{h}) + \alpha_n^{h}/N
\end{equation}

Finally $p^{\it M_c LS}_n$ or $p^{\it H_c LS}_n$ is used to calculate the cross-entropy loss $CE(p,\hat{p})^{LS}$ as in Equation~\eqref{eq:cels}:
\begin{equation}
\small
\label{eq:cemcls}
CE(p,\hat{p})^{\it M_c LS} = -\sum_{n=1}^{N}p^{\it M_c LS}_n\log(\hat{p}_n) 
\end{equation}
\begin{equation}
\small
\label{eq:cehcls}
CE(p,\hat{p})^{\it H_c LS} = -\sum_{n=1}^{N}p^{\it H_c LS}_n\log(\hat{p}_n) 
\end{equation}


\subsection{Model and Human Confidence Curriculum Learning}
We integrate the model and human confidence ($M_c$ and $H_c$) to set the ranking threshold $\mu$ to design the CL framework. Figure~\ref{fig:scoring ls} (right) demonstrates the CL training scheme with the model and human confidence as the scoring function. To design the CL training framework, $\mu$ is updated each epoch based on an update factor $\beta$ that controls the learning speed and decides the ratio of initial easy samples and ending epoch. We tune the initial rate of easier samples $r$ and ending epoch for CL training $e$ to determine the best ranking threshold $\mu$ and update factor $\beta$ as detailed below. In this section, we discuss how to set and update $\mu$ with the utilization of model and human confidence.

\paragraph{Model Confidence Curriculum Learning.}
Model confidence is obtained from the predicted probability for true classes by using an $iid$ baseline model. The higher model confidence means an easier sample. Based on this criteria, we choose easier samples (samples with higher confidence score) at the beginning of training then gradually using harder samples (samples with lower confidence scores) during loss calculation.
Suppose $M_c$ is the  model confidence for a sample (i.e. the predicted probability of the target label), then the cross-entropy loss with CL training $Loss^{CL}$ can be presented as follows:
\begin{equation}
\small
\label{eq:samples_mc}
Loss^{CL} =
\begin{cases}
CE(p,\hat{p}), \text { if } M_c \ge \mu \\ 
0, \text { otherwise}
\end{cases}
\end{equation}
In a mini-batch, harder samples with $M_c < \mu$ will be ignored in loss calculation. The value of $\mu$ can be reduced for successive batches progressively to implement CL.


\paragraph{Human Confidence Curriculum Learning.}
Human confidence is the multi-rater agreement distribution over target classes for a sample. If all annotators agree with one class for a sample, this sample is highly recognizable. On the contrary, if a sample is labeled with several classes, the sample is less identifiable as raters have different judgements over it. We compute the human confidence as $H_c = \sigma$, the standard derivation $\sigma$ of the distribution. A smaller $\sigma$ indicates that the probability mass is distributed more widely and the sample is harder. A higher $\sigma$ indicates an easier sample.
Given $H_c$ is the human confidence, the cross-entropy loss with CL training loss $Loss^{CL}$ can be represented as follows:

\begin{equation}
\small
\label{eq:samples_hc}
Loss^{CL} =
\begin{cases}
CE(p,\hat{p}), \text { if } H_c \ge \mu \\ 
0, \text { otherwise}
\end{cases}
\end{equation}

In a mini-batch, harder samples with $H_c < \mu$ will be ignored in loss calculation. The value of $\mu$ can be reduced for successive batches progressively to implement CL.

\paragraph{Update the Ranking Threshold $\mu$}
The ranking threshold $\mu$ is designed based on the initial ratio $r$ of easier samples for the training and ending epoch $e$ of CL. To fit the `easy to hard' training strategy in CL, we keep scaling $\mu$ so that the training subset gradually expands to use the entire training set. 
We update $\mu$ by scaling linearly using the dividing result of $r$ and the $e^{\it th}$:

\begin{align}
\small
\label{eq:beta}
\beta = \mu/e \\
\mu \leftarrow \mu - \beta
\end{align}

The full procedure of confidence-aware curriculum learning is summarized in Algorithm~\ref{alg:cl}. The proposed CL strategy is utilized to calculate the loss in mini-batch ($b$) and the ending of CL scheme is controlled by the update factor $\beta$. 
\begin{algorithm}[ht]
\caption{Confidence-Aware Curriculum Learning.}
\label{alg:cl}
\begin{algorithmic}[1]
\STATE{\textbf{Input:} Model parameter $w$, threshold $\mu$ determined from CL criteria, mini-batch size $b$.}
\STATE{\textbf{Set} $\beta$ by Eq.~\eqref{eq:beta}}
\WHILE{\textit{not converged}}
\FOR{$i=1$ to $b$}
\STATE{calculate loss via Eq.~\eqref{eq:samples_mc} or~\eqref{eq:samples_hc}}
\STATE{update w}
\ENDFOR
\ENDWHILE
\end{algorithmic}
\end{algorithm}
\section{Experimental Setup}
\label{sec:experiment}

\subsection{Datasets}
We report results on the main datasets we could find that included full-label distributions from multiple human annotators.

\paragraph{CIFAR10-H.} CIFAR10 is one of the benchmark datasets for image classification, which contains 10 classes in total, 50,000 images for training and 10,000 images for testing. CIFAR10-H~\cite{peterson2019human} includes the full-label distribution as annotated by humans only for each of the 10,000 images in the CIFAR10 test set. It utilizes around 500,000 crowdsourced human labels, with 50 annotators labelling each image on average. 
As we seek to use the human label distributions for training, we use these 10,000 images (CIFAR10-H test set) for training and report results instead on the CIFAR10-H training set of  50,000 images. In our experimental setup, the batch size was 1024 with a training set of 10,000 and test set of 50,000 images respectively.

\paragraph{WikiArt Emotions Dataset.} 
The WikiArt dataset~\cite{mohammad2018wikiart} reveals the interrelations between visual art, text describing it, and human emotion. It is relatively small, and consists of 4,105 pieces of art, selected from the WikiArt.org's collection with twenty-two categories (cubism, baroque etc). Each category covers about 200 items to make it a balanced dataset. Both the image of the art and its corresponding title are annotated through crowdsourcing independently using a 20-way classification of the emotion (contentment, amusement, sadness, etc.) evoked by the artwork or its title. Conceptually, the label distributions are different to those in CIFAR10-H, as artwork can evoke various emotional reactions in different people, while in CIFAR10-H, the distributions arise due to poor image quality.  Therefore the annotation distribution better reflects the affectual response to the image and text than a single label of the dominant class. As we had access to the larger CIFAR10-H dataset for the image classification task, we used the Wikiart Emotions Dataset only for text classification. The average length of the artwork titles was 5.8 words and each item was annotated 10 times on average. The batch size was 32 with 0.8 and 0.2 as the train and test split.

\subsection{Implementation Details}

To evaluate our method on an image classification task, we used the benchmark CIFAR10-H dataset. We report results using two state-of-the-art architectures: ResNet-34~\cite{he2016deep} with pretrained weights of ImageNet dataset~\cite{russakovsky2015imagenet}, and DenseNet-121~\cite{huang2017densely} with pretrained weights. We used the Stochastic Gradient Descent (SGD) optimizer with momentum as 0.9. The initial learning rate scheduler was 0.1 and it decays 0.1 for each 30 epochs.

To evaluate our method on a text classification task, we used the text component of the WikiArt dataset and report results using two models: BERT (Bidirectional Encoder Representations from Transformers)~\cite{devlin2018bert} and transformers~\cite{vaswani2017attention} adapted from Huggingface library~\cite{wolf2020transformers}. We chose AdamW~\cite{loshchilov2017decoupled} as the optimizer with learning rate 0.00002. 

To design a successful CL framework and select the best threshold $\mu$, we fine-tune the initial ratio $r$ of easier samples and ending epoch $e$ in all experiments. Based on our observation, the combination of a larger initial percentage of easy samples and an earlier ending epoch produces better accuracy in all experiments. It is consistent with the work of Gilmer et al.~\cite{gilmer2021loss}, which suggested that early stage initialization matters for deep neural networks. 


The GPU of Nvidia Tesla P40 with memory of 23GB was used for all experiments. For model and human confidence label smoothing ($M_c LS$/ $H_c LS$), the weighting parameter $\gamma$ was fine-tuned for each proposed method. 

\begin{table*}[]
\caption{Results of image and text classification tasks. Baselines for image classification task are vanilla model of ResNet34 and DenseNet121, and for text classification are BERT and transformers. In terms of training strategy, IID is plain training, McCL and HcCL are model and human confidence with CL. CE, LS, McLS and HcLS are cross-entropy loss, label smoothing, model and human confidence with LS respectively. The bold figures are the best results for each dataset. For Accuracy, higher values are superior, while for ECE, lower values are superior.}

\label{tab:main result}
\centering
\scalebox{0.9}{
\begin{tabular}{cccccccccc}
\hline
\multicolumn{1}{l|}{}                                                                             & \multicolumn{1}{l|}{}                      & \multicolumn{4}{c|}{CIFAR10-H}                                                                                                                                                        & \multicolumn{4}{c}{WikiArt}                                                                                                                                      \\ \hline
\multicolumn{1}{l|}{}                                                                             & \multicolumn{1}{l|}{}                      & \multicolumn{2}{c|}{ResNet34}                                                             & \multicolumn{2}{c|}{DenseNet121}                                                          & \multicolumn{2}{c|}{BERT}                                                                 & \multicolumn{2}{c}{Transformers}                                     \\ \hline
\multicolumn{1}{c|}{\multirow{2}{*}{\begin{tabular}[c]{@{}c@{}}Training\\ Strategy\end{tabular}}} & \multicolumn{1}{c|}{\multirow{2}{*}{Loss}} & \multicolumn{1}{c|}{\multirow{2}{*}{Acc(\%)$\uparrow$}} & \multicolumn{1}{c|}{\multirow{2}{*}{ECE$\downarrow$}} & \multicolumn{1}{c|}{\multirow{2}{*}{Acc(\%)$\uparrow$}} &  \multicolumn{1}{c|}{\multirow{2}{*}{ECE$\downarrow$}} & \multicolumn{1}{c|}{\multirow{2}{*}{Acc(\%)$\uparrow$}} & \multicolumn{1}{c|}{\multirow{2}{*}{ECE$\downarrow$}} & \multicolumn{1}{c|}{\multirow{2}{*}{Acc(\%)$\uparrow$}} & \multirow{2}{*}{ECE$\downarrow$} \\
\multicolumn{1}{c|}{}                                                                             & \multicolumn{1}{c|}{}                      & \multicolumn{1}{c|}{}                         & \multicolumn{1}{c|}{}                     & \multicolumn{1}{c|}{}                         & \multicolumn{1}{c|}{}                     & \multicolumn{1}{c|}{}                         & \multicolumn{1}{c|}{}                     & \multicolumn{1}{c|}{}                         &                      \\ \hline
\multicolumn{10}{c}{Baselines}                                                                                                                                                                                                                                                                                                                                                                                                                                                                            \\ \hline
\multicolumn{1}{c|}{IID}                                                                          & \multicolumn{1}{c|}{CE}                    & \multicolumn{1}{c|}{83.50}                    & \multicolumn{1}{c|}{0.1198}               & \multicolumn{1}{c|}{82.31}                    & \multicolumn{1}{c|}{0.1251}               & \multicolumn{1}{c|}{65.58}                    & \multicolumn{1}{c|}{0.1824}               & \multicolumn{1}{c|}{67.30}                    & 0.1644               \\ \hline
\multicolumn{1}{c|}{IID}                                                                          & \multicolumn{1}{c|}{LS}                    & \multicolumn{1}{c|}{84.78}                    & \multicolumn{1}{c|}{0.0705}               & \multicolumn{1}{c|}{82.81}                    & \multicolumn{1}{c|}{0.0548}               & \multicolumn{1}{c|}{66.26}                    & \multicolumn{1}{c|}{0.1651}               & \multicolumn{1}{c|}{68.11}                    & 0.1216               \\ \hline
\multicolumn{10}{c}{Proposed}                                                                                                                                                                                                                                                                                                                                                                                                                                                                             \\ \hline
\multicolumn{1}{c|}{IID}                                                                          & \multicolumn{1}{c|}{McLS}                  & \multicolumn{1}{c|}{85.37}                    & \multicolumn{1}{c|}{0.0557}               & \multicolumn{1}{c|}{83.16}                    & \multicolumn{1}{c|}{0.0437}               & \multicolumn{1}{c|}{66.62}                    & \multicolumn{1}{c|}{0.1782}               & \multicolumn{1}{c|}{68.81}                    & 0.1088               \\ \hline
\multicolumn{1}{c|}{IID}                                                                          & \multicolumn{1}{c|}{HcLS}                  & \multicolumn{1}{c|}{85.60}                    & \multicolumn{1}{c|}{0.0503}               & \multicolumn{1}{c|}{83.18}                    & \multicolumn{1}{c|}{0.0397}               & \multicolumn{1}{c|}{67.49}                    & \multicolumn{1}{c|}{0.1791}               & \multicolumn{1}{c|}{69.42}                    & 0.1036               \\ \hline
\multicolumn{1}{c|}{\multirow{2}{*}{McCL}}                                                        & \multicolumn{1}{c|}{McLS}                  & \multicolumn{1}{c|}{86.58}                    & \multicolumn{1}{c|}{0.0560}               & \multicolumn{1}{c|}{83.94}                    & \multicolumn{1}{c|}{0.0413}               & \multicolumn{1}{c|}{67.74}                    & \multicolumn{1}{c|}{\textbf{0.1481}}      & \multicolumn{1}{c|}{69.54}                    & 0.0986               \\ \cline{2-10} 
\multicolumn{1}{c|}{}                                                                             & \multicolumn{1}{c|}{HcLS}                  & \multicolumn{1}{c|}{86.18}                    & \multicolumn{1}{c|}{0.0623}               & \multicolumn{1}{c|}{83.81}                    & \multicolumn{1}{c|}{0.0521}               & \multicolumn{1}{c|}{68.12}                    & \multicolumn{1}{c|}{0.1561}               & \multicolumn{1}{c|}{69.30}                    & \textbf{0.0906}      \\ \hline
\multicolumn{1}{c|}{\multirow{2}{*}{HcCL}}                                                        & \multicolumn{1}{c|}{McLS}                  & \multicolumn{1}{c|}{86.19}                    & \multicolumn{1}{c|}{0.0543}               & \multicolumn{1}{c|}{83.74}                    & \multicolumn{1}{c|}{0.0556}               & \multicolumn{1}{c|}{68.01}                    & \multicolumn{1}{c|}{0.1623}               & \multicolumn{1}{c|}{70.02}                    & 0.1655               \\ \cline{2-10} 
\multicolumn{1}{c|}{}                                                                             & \multicolumn{1}{c|}{HcLS}                  & \multicolumn{1}{c|}{\textbf{86.81}}           & \multicolumn{1}{c|}{\textbf{0.0473}}      & \multicolumn{1}{c|}{\textbf{84.12}}           & \multicolumn{1}{c|}{\textbf{0.0360}}      & \multicolumn{1}{c|}{\textbf{68.45}}           & \multicolumn{1}{c|}{0.1534}               & \multicolumn{1}{c|}{\textbf{71.15}}           & 0.0918               \\ \hline
\end{tabular}
}
\end{table*}

\section{Results}
\label{sec:result}
We report the `top-1' accuracy to measure the model performance and Expected Calibration Error (ECE)~\cite{naeini2015obtaining} as the primary metric for calibration. ECE divides predictions into $M$ equally-spaced bins and takes the weighted mean of each bin's confidence gap. Given $B_{m}$ is the set of indices of samples, $acc(B_{m})$ and $conf(B_{m})$ are the average accuracy and confidence of each bin and $n$ is the sample size, the equation of ECE is: 
\begin{equation}
\small
\label{eq:ece}
\mathrm{ECE}=\sum_{m=1}^{M} \frac{\left|B_{m}\right|}{n}\left|\operatorname{acc}\left(B_{m}\right)-\operatorname{conf}\left(B_{m}\right)\right|.
\end{equation}

We choose $M=15$ bins in all experiments with the reference of the work in Guo \emph{et al}~\cite{guo2017calibration}. 

\subsection{Image Classification}
The left columns in Table~\ref{tab:main result} shows the results for image classification using the CIFAR10-H dataset. With ResNet-34 and independent identically distributed ($iid$) training, both model and human confidence label smoothing ($HCLS$ and $MCLS$) outperform baseline and uniform label smoothing in terms of accuracy and calibration. The CL training strategy outperforms the $iid$ strategy, and the model using human confidence for LS and for ranking with CL has the best accuracy and calibration. With DenseNet-121 architecture, the best accuracy and ECE are also reported for the $HCCL$ training strategy and the trends are similar to ResNet-34. 

In summary, our proposed methods show promising results for both accuracy and ECE, as seen in the bold cells in Table~\ref{tab:main result}. Models trained with proposed methods are better calibrated than the baselines and achieve better accuracy, especially when used in a curriculum learning framework. 
This can also be seen in the reliability diagram in Figure~\ref{fig:rd} (left), where both $MCLS$ and $HCLS$ are nearer to the perfect calibration line than the baselines.  

\subsection{Text Classification}
The results for text classification using the WikiArt dataset are presented in the right of Table~\ref{tab:main result}. With the $iid$ training for BERT model, the proposed $MCLS$ and $HCCL$ produce better accuracy than baseline CE and LS. In terms of curriculum learning, the combination of $HCCL$ and $HCLS$ obtains the best accuracy. Curriculum learning again has the effect of reducing ECE compared to $iid$ training, with $MCLS$ performing best. 

With the transformer architecture, trends are similar with the best accuracy again obtained with $HCCL$ using the $HCLS$ loss function. There is more variation with the ECE metric, but all the proposed methods perform better than the two baselines.

To summarise, as for the image dataset, the curriculum learning strategies outperform all the $iid$ strategies for text classification. Compared to the CIFAR10H dataset, improvements in calibration in particular are less consistent for the WikiArt dataset. This is largely due to the small size of the dataset and the greater sophistication of the task, with more and harder to distinguish labels. 

\begin{figure}[!h]
\centerline{\includegraphics[width=0.8\textwidth]{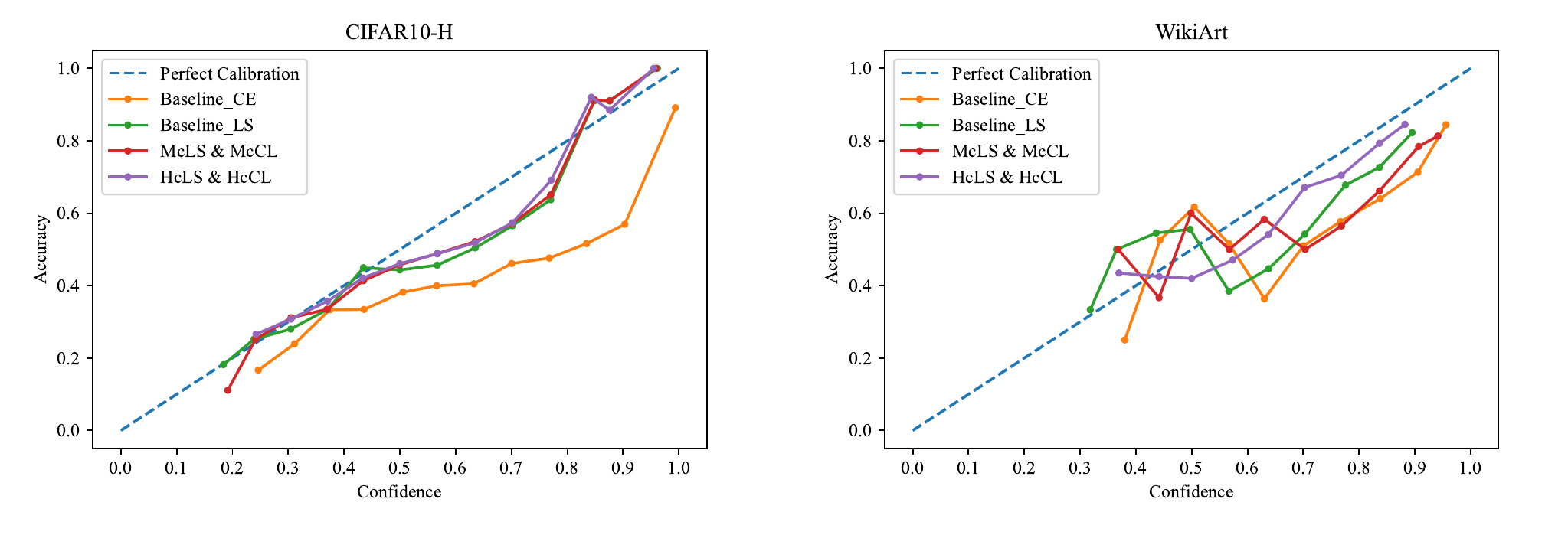}}
\caption{Reliability diagram for image and text classification with ResNet34 and transformer respectively. $HcLS\&HcCL$ denotes the human confidence label smoothing as loss function in CL with human confidence to rank samples. $McLS\&McCL$ represents the CL strategy with model confidence to set difficulty score for samples and model confidence label smoothing as loss function.}
\label{fig:rd}
\end{figure}

This can be seen graphically in the reliability diagram in Figure~\ref{fig:rd} (right), where all the models deviate substantially from the perfect calibration line, though the model with $HCCL$ using the $HCLS$ loss function is closer to the perfect calibration than the rest. 

\section{Conclusion}

In this paper, we used notions of model and human confidence to improve upon standard Label Smoothing, where a proportion of probability mass is uniformly distributed from the one-hot label to the other labels. We then demonstrated that these can be effectively used as the scoring function in curriculum learning. Our confidence-aware approaches outperform baselines in terms of accuracy and ECE for image and text classification and across a range of neural architectures, as shown in Table~\ref{tab:main result}. The training strategies using human confidence as the ranking criteria overall obtain better accuracy and ECE than those using model confidence, indicating the superiority of human perception in guiding the design of the curriculum learning framework. However, as human confidence is expensive and time-consuming to collect, open-source datasets with multi-rater labels are very hard to obtain, which might diminish the benefits of those proposed methods. On the other hand, the performance of the learning strategy using model confidence also outperforms the baselines. As the model confidence is easier to collect, just requiring the pre-running of a baseline $iid$ classifier, we can utilize the model confidence for many more datasets, and also apply such methods to other tasks than classification, such as segmentation and detection. Still, some human tasks are genuinely subjective, and it is important that the range of human perceptions to a sample are captured. Emotional response to art is one such example, and in future work we would like to expand the size of datasets such as WikiArt. We also wish to explore other possibilites for ranking training samples in Curriculum Learning from emotion datasets, for example utilising the intensity and valence of emotions.

\bibliography{bib} 

\begin{thebibliography}{10}

\bibitem{szegedy2016rethinking}
C. Szegedy, V. Vanhoucke, S. Ioffe, J. Shlens and Z. Wojna, ``Rethinking the
  inception architecture for computer vision,'' in [{\em Proceedings of the
  IEEE conference on computer vision and pattern
  recognition}{\nolinebreak\hspace{0.1em}]},   2818--2826 (2016).

\bibitem{muller2019does}
R. M{\"u}ller, S. Kornblith and G.~E. Hinton, ``When does label smoothing
  help?,'' {\em Advances in neural information processing systems}~{\bf 32}
  (2019).

\bibitem{peterson2019human}
J.~C. Peterson, R.~M. Battleday, T.~L. Griffiths and O. Russakovsky, ``Human
  uncertainty makes classification more robust,'' in [{\em Proceedings of the
  IEEE/CVF International Conference on Computer
  Vision}{\nolinebreak\hspace{0.1em}]},   9617--9626 (2019).

\bibitem{bengio2009curriculum}
Y. Bengio, J. Louradour, R. Collobert and J. Weston, ``Curriculum learning,''
  in [{\em Proceedings of the 26th annual international conference on machine
  learning}{\nolinebreak\hspace{0.1em}]},   41--48 (2009).

\bibitem{guo2018curriculumnet}
S. Guo, W. Huang, H. Zhang, C. Zhuang, D. Dong, M.~R. Scott and D. Huang,
  ``Curriculumnet: Weakly supervised learning from large-scale web images,'' in
  [{\em Proceedings of the European Conference on Computer Vision
  (ECCV)}{\nolinebreak\hspace{0.1em}]},   135--150 (2018).

\bibitem{wei2021learn}
J. Wei, A. Suriawinata, B. Ren, X. Liu, M. Lisovsky, L. Vaickus, C. Brown, M.
  Baker, M. Nasir-Moin, N. Tomita et~al., ``Learn like a pathologist:
  curriculum learning by annotator agreement for histopathology image
  classification,'' in [{\em Proceedings of the IEEE/CVF Winter Conference on
  Applications of Computer Vision}{\nolinebreak\hspace{0.1em}]},   2473--2483
  (2021).

\bibitem{xu2020curriculum}
B. Xu, L. Zhang, Z. Mao, Q. Wang, H. Xie and Y. Zhang, ``Curriculum learning
  for natural language understanding,'' in [{\em Proceedings of the 58th Annual
  Meeting of the Association for Computational
  Linguistics}{\nolinebreak\hspace{0.1em}]},   6095--6104 (2020).

\bibitem{tay2019simple}
Y. Tay, S. Wang, L.~A. Tuan, J. Fu, M.~C. Phan, X. Yuan, J. Rao, S.~C. Hui and
  A. Zhang, ``Simple and effective curriculum pointer-generator networks for
  reading comprehension over long narratives,'' {\em arXiv preprint
  arXiv:1905.10847}  (2019).

\bibitem{ren2018self}
Z. Ren, D. Dong, H. Li and C. Chen, ``Self-paced prioritized curriculum
  learning with coverage penalty in deep reinforcement learning,'' {\em IEEE
  transactions on neural networks and learning systems}~{\bf 29}(6),
  2216--2226 (2018).

\bibitem{platanios2019competence}
E.~A. Platanios, O. Stretcu, G. Neubig, B. Poczos and T.~M. Mitchell,
  ``Competence-based curriculum learning for neural machine translation,'' {\em
  arXiv preprint arXiv:1903.09848}  (2019).

\bibitem{kocmi2017curriculum}
T. Kocmi and O. Bojar, ``Curriculum learning and minibatch bucketing in neural
  machine translation,'' {\em arXiv preprint arXiv:1707.09533}  (2017).

\bibitem{wei2016stc}
Y. Wei, X. Liang, Y. Chen, X. Shen, M.-M. Cheng, J. Feng, Y. Zhao and S. Yan,
  ``Stc: A simple to complex framework for weakly-supervised semantic
  segmentation,'' {\em IEEE transactions on pattern analysis and machine
  intelligence}~{\bf 39}(11),  2314--2320 (2016).

\bibitem{qin2020balanced}
W. Qin, Z. Hu, X. Liu, W. Fu, J. He and R. Hong, ``The balanced loss curriculum
  learning,'' {\em IEEE Access}~{\bf 8},  25990--26001 (2020).

\bibitem{penha2020curriculum}
G. Penha and C. Hauff, ``Curriculum learning strategies for ir,'' in [{\em
  European Conference on Information Retrieval}{\nolinebreak\hspace{0.1em}]},
  699--713, Springer (2020).

\bibitem{zhang2018empirical}
X. Zhang, G. Kumar, H. Khayrallah, K. Murray, J. Gwinnup, M.~J. Martindale, P.
  McNamee, K. Duh and M. Carpuat, ``An empirical exploration of curriculum
  learning for neural machine translation,'' {\em arXiv preprint
  arXiv:1811.00739}  (2018).

\bibitem{everinghampascal}
M. Everingham, L. Van~Gool, C. Williams, J. Winn and A. Zisserman, ``The pascal
  visual object classes challenge 2012 results, vol. 5 (2012).''

\bibitem{tudor2016hard}
R. Tudor~Ionescu, B. Alexe, M. Leordeanu, M. Popescu, D.~P. Papadopoulos and V.
  Ferrari, ``How hard can it be? estimating the difficulty of visual search in
  an image,'' in [{\em Proceedings of the IEEE Conference on Computer Vision
  and Pattern Recognition}{\nolinebreak\hspace{0.1em}]},   2157--2166 (2016).

\bibitem{pentina2015curriculum}
A. Pentina, V. Sharmanska and C.~H. Lampert, ``Curriculum learning of multiple
  tasks,'' in [{\em Proceedings of the IEEE Conference on Computer Vision and
  Pattern Recognition}{\nolinebreak\hspace{0.1em}]},   5492--5500 (2015).

\bibitem{zoph2018learning}
B. Zoph, V. Vasudevan, J. Shlens and Q.~V. Le, ``Learning transferable
  architectures for scalable image recognition,'' in [{\em Proceedings of the
  IEEE conference on computer vision and pattern
  recognition}{\nolinebreak\hspace{0.1em}]},   8697--8710 (2018).

\bibitem{chorowski2016towards}
J. Chorowski and N. Jaitly, ``Towards better decoding and language model
  integration in sequence to sequence models,'' {\em arXiv preprint
  arXiv:1612.02695}  (2016).

\bibitem{vaswani2017attention}
A. Vaswani, N. Shazeer, N. Parmar, J. Uszkoreit, L. Jones, A.~N. Gomez, {\L}.
  Kaiser and I. Polosukhin, ``Attention is all you need,'' {\em Advances in
  neural information processing systems}~{\bf 30} (2017).

\bibitem{zhang2021understanding}
C. Zhang, S. Bengio, M. Hardt, B. Recht and O. Vinyals, ``Understanding deep
  learning (still) requires rethinking generalization,'' {\em Communications of
  the ACM}~{\bf 64}(3),  107--115 (2021).

\bibitem{lukasik2020does}
M. Lukasik, S. Bhojanapalli, A. Menon and S. Kumar, ``Does label smoothing
  mitigate label noise?,'' in [{\em International Conference on Machine
  Learning}{\nolinebreak\hspace{0.1em}]},   6448--6458, PMLR (2020).

\bibitem{korbar2018cooperative}
B. Korbar, D. Tran and L. Torresani, ``Cooperative learning of audio and video
  models from self-supervised synchronization,'' {\em Advances in Neural
  Information Processing Systems}~{\bf 31} (2018).

\bibitem{ghasedi2019balanced}
K. Ghasedi, X. Wang, C. Deng and H. Huang, ``Balanced self-paced learning for
  generative adversarial clustering network,'' in [{\em Proceedings of the
  IEEE/CVF Conference on Computer Vision and Pattern
  Recognition}{\nolinebreak\hspace{0.1em}]},   4391--4400 (2019).

\bibitem{buhrmester2016amazon}
M. Buhrmester, T. Kwang and S.~D. Gosling, ``Amazon's mechanical turk: A new
  source of inexpensive, yet high-quality data?,'' (2016).

\bibitem{achlioptas2021artemis}
P. Achlioptas, M. Ovsjanikov, K. Haydarov, M. Elhoseiny and L.~J. Guibas,
  ``Artemis: Affective language for visual art,'' in [{\em Proceedings of the
  IEEE/CVF Conference on Computer Vision and Pattern
  Recognition}{\nolinebreak\hspace{0.1em}]},   11569--11579 (2021).

\bibitem{mohammad2018wikiart}
S. Mohammad and S. Kiritchenko, ``Wikiart emotions: An annotated dataset of
  emotions evoked by art,'' in [{\em Proceedings of the eleventh international
  conference on language resources and evaluation (LREC
  2018)}{\nolinebreak\hspace{0.1em}]},  (2018).

\bibitem{he2016deep}
K. He, X. Zhang, S. Ren and J. Sun, ``Deep residual learning for image
  recognition,'' in [{\em Proceedings of the IEEE conference on computer vision
  and pattern recognition}{\nolinebreak\hspace{0.1em}]},   770--778 (2016).

\bibitem{russakovsky2015imagenet}
O. Russakovsky, J. Deng, H. Su, J. Krause, S. Satheesh, S. Ma, Z. Huang, A.
  Karpathy, A. Khosla, M. Bernstein et~al., ``Imagenet large scale visual
  recognition challenge,'' {\em International journal of computer vision}~{\bf
  115}(3),  211--252 (2015).

\bibitem{huang2017densely}
G. Huang, Z. Liu, L. Van Der~Maaten and K.~Q. Weinberger, ``Densely connected
  convolutional networks,'' in [{\em Proceedings of the IEEE conference on
  computer vision and pattern recognition}{\nolinebreak\hspace{0.1em}]},
  4700--4708 (2017).

\bibitem{devlin2018bert}
J. Devlin, M.-W. Chang, K. Lee and K. Toutanova, ``Bert: Pre-training of deep
  bidirectional transformers for language understanding,'' {\em arXiv preprint
  arXiv:1810.04805}  (2018).

\bibitem{wolf2020transformers}
T. Wolf, L. Debut, V. Sanh, J. Chaumond, C. Delangue, A. Moi, P. Cistac, T.
  Rault, R. Louf, M. Funtowicz et~al., ``Transformers: State-of-the-art natural
  language processing,'' in [{\em Proceedings of the 2020 conference on
  empirical methods in natural language processing: system
  demonstrations}{\nolinebreak\hspace{0.1em}]},   38--45 (2020).

\bibitem{loshchilov2017decoupled}
I. Loshchilov and F. Hutter, ``Decoupled weight decay regularization,'' {\em
  arXiv preprint arXiv:1711.05101}  (2017).

\bibitem{gilmer2021loss}
J. Gilmer, B. Ghorbani, A. Garg, S. Kudugunta, B. Neyshabur, D. Cardoze, G.
  Dahl, Z. Nado and O. Firat, ``A loss curvature perspective on training
  instability in deep learning,'' {\em arXiv preprint arXiv:2110.04369}
  (2021).

\bibitem{naeini2015obtaining}
M.~P. Naeini, G. Cooper and M. Hauskrecht, ``Obtaining well calibrated
  probabilities using bayesian binning,'' in [{\em Twenty-Ninth AAAI Conference
  on Artificial Intelligence}{\nolinebreak\hspace{0.1em}]},  (2015).

\bibitem{guo2017calibration}
C. Guo, G. Pleiss, Y. Sun and K.~Q. Weinberger, ``On calibration of modern
  neural networks,'' in [{\em International conference on machine
  learning}{\nolinebreak\hspace{0.1em}]},   1321--1330, PMLR (2017).

\end{thebibliography}
\bibliographystyle{spiebib} 

\end{document}